\documentclass{article}

\usepackage[preprint]{neurips_2024}  % Official NeurIPS style file (download from conference site)

\usepackage{amsmath}
\usepackage{amssymb}
\usepackage{amsthm}
\usepackage{graphicx}
\usepackage{hyperref}
\usepackage{url}
\usepackage{booktabs}  % For better tables
\usepackage{microtype}  % Improved typography
\newcounter{algocounter}

\title{Frequency-Forcing: From Scaling-as-Time to Soft Frequency Guidance}

\author{
  Weitao Du \\
  DAMO Academy \\
  duweitao.dwt@alibaba-inc.com
}

\begin{document}

\maketitle

\begin{abstract}
While standard flow-matching models transport noise to data uniformly along a single timeline, incorporating an explicit \textit{generation order}---specifically, establishing coarse, low-frequency structure before fine, high-frequency detail---has proven highly effective for synthesizing natural images. Two recent works offer distinct paradigms for enforcing such ordering. K-Flow \cite{du2025flow} imposes a \textit{hard} frequency constraint by reinterpreting a frequency scaling variable as the flow time, running the trajectory entirely inside a transformed $K$-amplitude space. Latent Forcing \cite{baade2026latent} provides a \textit{soft} ordering mechanism by coupling the pixel flow with an auxiliary semantic latent flow via asynchronous time schedules, leaving the pixel interpolation path itself untouched. Viewed purely from the angle of improving the pixel generation task, we observe that \textit{forcing}---guiding generation with an earlier-maturing auxiliary stream---offers a highly compatible and flexible route to scale-ordered generation without rewriting the core flow coordinate. Building on this observation, we propose \textbf{Frequency-Forcing}, which realizes K-Flow's frequency ordering through Latent Forcing's soft mechanism: a standard pixel flow is guided by an auxiliary low-frequency stream that matures earlier in time. Unlike Latent Forcing, whose semantic scratchpad relies on a heavy pretrained encoder (e.g., DINO), our frequency scratchpad is derived from the data itself via a lightweight learnable wavelet packet transform. We term this a \textit{self-forcing} signal, which avoids external dependencies while learning a basis better adapted to data statistics than the fixed bases (e.g., Haar) used in hard frequency flows. On ImageNet-256, Frequency-Forcing consistently improves FID over strong pixel- and latent-space baselines, and naturally composes with a semantic stream to yield further gains. This illustrates that forcing-based scale ordering is a versatile, path-preserving alternative to hard frequency flows.
\end{abstract}

\section{Introduction}
\label{sec:introduction}

Good generation proceeds from high-level structure to low-level detail: architects sketch massing before facades, directors storyboard before shooting, and natural images are best denoised from coarse, low-frequency content to fine, high-frequency detail \cite{dieleman2024spectral}. This ordering is grounded in the learning dynamics of neural networks: \textit{spectral bias} \cite{rahaman2019spectral} and the inverse-variance spectral law of diffusion trajectories predict that high-variance low-frequency modes are mastered orders of magnitude faster than fine details. Standard flow-matching models \cite{lipman2022flow} do pick up this coarse-to-fine ordering \emph{implicitly}, simply because low frequencies are easier to fit during high noise regime; but because all frequencies are denoised under a single synchronous clock, the ordering is noisy, entangled with gradient competition across bands, and offers no explicit knob for when structure is committed. Empirically, as our ablations confirm, this implicit ordering is weaker than \emph{explicit} frequency guidance---injecting a low-frequency signal into the generation path gives a clean gain on top of plain pixel-space flow matching.

How should explicit ordering be injected? Two recent works offer distinct paradigms. \textbf{K-Flow} \cite{du2025flow} formalizes explicit scale ordering \emph{directly along the frequency axis}: it reinterprets a scaling variable $k$ indexing a frequency (e.g., wavelet or Fourier) decomposition as the flow time, and runs flow matching inside the resulting transformed $K$-amplitude space. Generation then proceeds from low-$k$ to high-$k$ by construction, which in turn enables transformed-space capabilities like unsupervised frequency editing and training-free frequency-domain restoration. The price on the generative side is structural: endpoints and intermediate states are redefined in a transformed space, so (i) the vector field is anchored to a specific pre-chosen basis (Haar, Daubechies, PCA, Fourier\ldots) rather than one adapted to data statistics, and (ii) the trajectory is no longer the standard linear flow-matching interpolation, making the model incompatible with vanilla flow-matching pipelines and checkpoints. We call this regime a \textit{hard} frequency constraint---ordering is achieved by \textit{rewriting the flow coordinate}. Notably, K-Flow's original taxonomy of ``hard'' versus ``soft'' bumping cleanly anticipates and encompasses the various time-scheduling mechanisms proposed by later methods like Latent Forcing, and we use it as a unified vocabulary to frame our discussion of generation order (Table~\ref{tab:kflow_latentforcing_similarity}).

\textbf{Latent Forcing} \cite{baade2026latent} orders generation along a \emph{different axis}---high-level semantics before raw pixels---with a forcing regularization mechanism. It keeps the pixel interpolation end-to-end linear and instead couples it with an auxiliary \emph{semantic} latent flow via a pair of asynchronous time variables, scheduling the semantic stream to mature \textit{earlier} so that it acts as a ``scratchpad'' softly conditioning pixel denoising. The key property is that this \textit{forcing} formulation decouples \textit{ordering} (which representation matures first) from \textit{interpolation} (what the pixel flow is solving), making the mechanism agnostic to whether the auxiliary stream carries semantic, structural, or other information. In practice, however, Latent Forcing instantiates the scratchpad with a \textit{pretrained self-supervised encoder} (e.g., DINO \cite{caron2021dino}), effectively outsourcing the earlier-stage workspace to a heavy external model and not exploiting the low-to-high frequency ordering intrinsic to images.

\textbf{Our perspective: use the flexibility of forcing to inject a frequency scratchpad, self-sourced from the data.} K-Flow's choice of ordering axis (frequency) and Latent Forcing's \emph{mechanism} (asynchronous multi-time forcing) are separable design dimensions. For the specific goal of \emph{leveraging frequency structure to improve generation quality}, forcing is the more flexible of the two: any earlier-maturing auxiliary representation can be plugged in without touching the pixel interpolation, and the frequency prior can be scaled or composed with other priors without redesigning the trajectory. (We do not claim to subsume K-Flow's transformed-space tasks such as unsupervised frequency editing, which remain natural strengths of the hard-constraint formulation.) Taking this viewpoint, we propose \textbf{Frequency-Forcing}: we keep Latent Forcing's multi-time forcing formulation but replace its pretrained semantic scratchpad with a \textit{low-frequency} scratchpad derived from the data itself via a lightweight wavelet packet transform---no DINO-size pretrained encoder, and target extraction requires minimal additional supervision. We call this a \textit{self-forcing} signal, in contrast to Latent Forcing's \textit{semantic forcing} via distilled external representations.

Two further design choices make Frequency-Forcing a clean instance of this self-forcing idea. \emph{(1) Learnable wavelet basis.} K-Flow and related frequency-aware methods fix the basis a priori (Haar, Daubechies, Laplacian, \ldots), so the low-frequency scratchpad is determined more by a design choice than by the data. We instead learn the analysis filter from data via sparsity and quadrature-mirror-filter regularizers, yielding a sparse, approximately orthogonal wavelet-packet basis that outperforms fixed Haar or Laplacian bases. \emph{(2) Path-preserving architecture.} Since the pixel branch remains a standard linear flow-matching interpolation, Frequency-Forcing is structurally aligned with vanilla flow matching and---because forcing is attached as auxiliary streams rather than baked into the trajectory---composes with additional priors out of the box: as a scalability bonus, adding a DINO stream on top of the frequency stream yields further gains (Section~\ref{sec:experiments}), a composition that would be hard to realize in K-Flow's transformed-space setting. Fine-tuning a pretrained pixel checkpoint with an attached frequency stream is a natural next step that we leave to future work.

\subsection{Contributions}
The main contributions of this work are:
\begin{itemize}
    \item \textbf{A forcing-based realization of scale-ordered generation.} We identify that the \textit{forcing} formulation of Latent Forcing \cite{baade2026latent} provides a flexible, path-preserving alternative to K-Flow-style frequency-coordinate rewriting \cite{du2025flow}. Concretized on the frequency axis, this gives a new method---\textbf{Frequency-Forcing}---that realizes coarse-to-fine generation as soft, asynchronous conditioning on top of a standard flow-matching pixel trajectory.
    \item \textbf{Self-forcing vs.\ semantic forcing.} Unlike Latent Forcing, whose scratchpad is a pretrained self-supervised encoder (e.g., DINO), our auxiliary stream is derived from the data itself via a cheap low-frequency projection. We argue, and empirically validate, that this \textit{self-forcing} view makes the forcing paradigm usable in settings where strong pretrained encoders are unavailable or undesirable.
    \item \textbf{Learnable, data-adapted wavelet basis.} Instead of relying on a fixed, pre-chosen frequency basis as in hard frequency flows, we learn a sparse, approximately orthogonal wavelet-packet basis from the dataset itself and show it outperforms fixed Haar and Laplacian bases.
    \item \textbf{Extensibility and empirical validation.} Because forcing is attached as auxiliary streams, Frequency-Forcing composes with additional priors out of the box: adding a DINO stream on top yields further gains, illustrating its compositional advantage over transformed-space methods. On ImageNet-256, Frequency-Forcing consistently improves FID over strong pixel-space and latent-space baselines.
\end{itemize}

\section{Related Work}
\label{sec:related}

\paragraph{Frequency- and Scale-Ordered Generative Modeling.}
Continuous-time generative models \cite{ho2020denoising,lipman2022flow} transport a simple prior to a complex data distribution, but their standard trajectories are blind to the frequency and scale hierarchies of images, and a long line of work makes this hierarchy explicit. Laplacian-pyramid GANs \cite{denton2015lapgan} and cascaded diffusion \cite{ho2022cascaded} train separate models per resolution and stitch them by renoising/upsampling. Scale-autoregressive generators instead predict the \emph{next scale} rather than the next token: VAR \cite{tian2024var} realizes this for discrete tokens, and FlowAR \cite{ren2024flowar} combines scale-wise autoregression with per-scale flow matching. K-Flow \cite{du2025flow}, which is closest to us in motivation, takes the continuous version of the same idea and rewrites the entire flow trajectory in a transformed frequency domain; as discussed in Section~\ref{sec:introduction}, this yields a \textit{hard} frequency constraint---a fixed basis and a non-linear trajectory that breaks compatibility with vanilla flow-matching checkpoints. Frequency-Forcing shares the coarse-to-fine intuition of all these methods, but avoids cascades, scale autoregression, and trajectory rewriting: a single backbone runs a linear pixel flow alongside an auxiliary frequency stream whose coarse-to-fine behavior arises purely from asynchronous per-stream schedules.

\paragraph{Pixel-Space and Representation-Assisted Generation.}
A second axis along which modern generative modeling is organized is \emph{where} the diffusion process is run. Latent diffusion models \cite{rombach2022ldm} denoise inside a pretrained VAE latent and underpin most state-of-the-art systems, but pay a reconstruction--generation price: the encoder is lossy and its latent statistics are decoupled from generator design. Pixel-space diffusion has recently closed much of this gap---JiT \cite{li2025jit} restores pixel-space DiTs to competitive FID with careful loss design, while spectral analyses \cite{dieleman2024spectral} show that pixel training already implicitly performs a coarse-to-fine spectral autoregression. Orthogonally, a fast-growing line of work argues that the main value of latent diffusion is not compression but the use of \emph{understanding-oriented representations} to guide generation: REPA \cite{yu2024repa} distills features from self-supervised encoders (e.g., DINOv2 \cite{oquab2023dinov2}) into a standard DiT, VA-VAE \cite{yao2025vavae} aligns tokenizer latents with vision features, and \emph{representation autoencoders} (RAE \cite{zheng2025rae}, REPA-E \cite{leng2025repae}, SVG \cite{shi2025svg}) replace the reconstruction-only VAE with a frozen or end-to-end-tuned representation encoder, showing that semantically rich latents substantially accelerate convergence and improve fidelity. Latent Forcing \cite{baade2026latent} reframes this line as a matter of \emph{generation order}: it keeps a pixel flow but couples it with a DINO latent that denoises earlier in time, so that the ``understanding'' signal conditions pixel generation without being baked into a lossy tokenizer. Frequency-Forcing adopts the same multi-time forcing formulation but observes that the scratchpad need not be a heavy semantic encoder: an image-intrinsic, self-sourced low-frequency latent already serves as a strong earlier-maturing stream, and the two priors compose naturally (Sec.~\ref{sec:method_extensibility}).

\section{Frequency-Forcing}
\label{sec:method}

\subsection{Soft Frequency Guidance via an Auxiliary Stream}
\label{sec:method_formulation}
Standard flow matching defines a vector field that transports a noise distribution to a data distribution over a single global time variable $t \in [0,1]$, with $z(t) = t\,x + (1-t)\epsilon$. K-Flow modifies this interpolation path: 
the target $x$ and intermediate states are replaced by their transformed-domain counterparts so that the flow evolves along the frequency axis. As discussed in Section~\ref{sec:introduction}, this tightly couples the pixel trajectory to the chosen frequency basis and breaks compatibility with vanilla flow matching.

\textbf{Frequency-Forcing} instead preserves the pixel \emph{interpolation path} (still a linear noise-to-data flow) and adds a \textit{parallel auxiliary stream} that carries a low-frequency latent. The frequency stream does influence the \emph{learned} pixel vector field---via the shared backbone, it acts as an implicit regularizer and a soft conditioning signal---but it does so without redefining the pixel trajectory itself. Concretely, given data $x$ and a frequency extractor $\mathcal{F}$ (defined in Sec.~\ref{sec:learnable_wavelet}), we run two coupled flow-matching processes:
\begin{equation}
z_\text{pix}(t_\text{pix}) = t_\text{pix}\, x + (1-t_\text{pix})\,\epsilon_\text{pix}, \qquad z_\text{fre}(t_\text{fre}) = t_\text{fre}\, x_\text{fre} + (1-t_\text{fre})\,\epsilon_\text{fre},
\label{eq:two_stream}
\end{equation}
where $x_\text{fre} = \mathcal{F}(x)$ represents the low-frequency latent (extracted via low-pass wavelet reconstruction followed by patch embedding, see Sec.~\ref{sec:learnable_wavelet}), $\epsilon_\text{pix},\epsilon_\text{fre}\sim\mathcal{N}(0,I)$, and the two clocks $t_\text{pix}, t_\text{fre}$ are tied by $t_\text{fre} = \phi(t_\text{pix})$ with $\phi$ chosen so that the frequency stream \textit{matures earlier} than the pixel stream along the trajectory. This asynchronous, earlier-maturation schedule directly inherits the \textit{forcing} formulation of Latent Forcing \cite{baade2026latent}; the only difference is that the earlier-maturing workspace is now an explicit frequency latent rather than a semantic one. Crucially, the pixel branch in Eq.~\eqref{eq:two_stream} is identical to a standard linear flow-matching interpolation---only the \textit{conditioning signal seen by the backbone} is augmented by the partially denoised $z_\text{fre}$.

\paragraph{Soft vs.\ hard frequency constraints.} We call this a \textit{soft} frequency constraint: the frequency prior biases the pixel vector field only through the shared backbone, without redefining its endpoints or trajectory. Removing the auxiliary stream recovers vanilla flow matching, and the guidance strength varies smoothly with how much earlier the frequency stream matures. K-Flow's \textit{hard} constraint sits on a qualitatively different trajectory---its flow time is a frequency scaling variable and its endpoints are transformed-domain objects---so the two formulations are not connected by a continuous limit; Frequency-Forcing, by design, keeps the pixel trajectory structurally identical to standard flow matching while still delivering a coarse-to-fine structural benefit. It is also worth noting that K-Flow's original \emph{hard-}/\emph{soft-}bumping taxonomy---which describes how sharply consecutive frequency bands are activated along K-Flow's own trajectory---turns out to naturally cover the different time schedulers used by Latent Forcing as well, and we use this shared vocabulary to locate all three methods on the same axis (Table~\ref{tab:kflow_latentforcing_similarity}).

\paragraph{Training objective.} Throughout the paper we adopt the same $x$-prediction training objective as \cite{baade2026latent}, in which the network directly predicts the clean target $x^{(m)}$ of each stream $m \in \{\text{pix}, \text{fre}\}$ from the noised state $z^{(m)}_{t_m}$ produced by Eq.~\eqref{eq:two_stream}. Let $x_{\text{pred}}^{(m)}(z^{(m)}_{t_m}, t_m)$ denote this prediction. The per-stream loss uses velocity-matching weighting:
\begin{equation}
\mathcal{L} = \sum_{m} \lambda_{m}\mathcal{L}_{m}, \qquad
\mathcal{L}_m = \left\| \frac{x_{\text{pred}}^{(m)} - z^{(m)}_{t_m}}{\max(1-t_m, t_{\text{clip}})} - \frac{x^{(m)} - z^{(m)}_{t_m}}{\max(1-t_m, t_{\text{clip}})} \right\|^2,
\label{eq:x_pred_loss}
\end{equation}
where $z^{(m)}_{t_m} = t_m x^{(m)} + (1-t_m)\epsilon^{(m)}$ is the noised sample from Eq.~\eqref{eq:two_stream}, $\epsilon^{(m)}\!\sim\!\mathcal{N}(0,I)$ is the sampled noise, $\lambda_m$ balances the two streams, and $t_{\text{clip}}$ is a small positive constant (we use $t_{\text{clip}}=0.05$ following \cite{li2025jit}) that prevents $1{-}t_m$ from vanishing and causing numerical blow-up near 
$t_m\!\to\!1$. This objective corresponds to predicting the clean target $x^{(m)}$ under an $x$-prediction parameterization, while matching the induced velocity target $(x^{(m)} - z^{(m)}_{t_m})/(1-t_m)$. With clipping near $t_m \to 1$, it becomes a numerically stabilized approximation to the standard velocity-matching loss.
%On the pixel stream this reduces to exactly the vanilla flow-matching loss, which is why the pixel trajectory remains unchanged and the method is, at the level of architecture and loss, structurally aligned with vanilla flow matching.

\begin{table}[ht]
\centering
\small
\begin{tabular}{p{0.22\textwidth} p{0.35\textwidth} p{0.35\textwidth}}
\toprule
\textbf{Aspect} & \textbf{K-Flow} \cite{du2025flow} & \textbf{Latent Forcing} \cite{baade2026latent} \\
\midrule
Flow-time parametrization
& Time is the scaling variable $k$ indexing a frequency basis.
& Separate time variables are used for the latent and pixel streams. \\

Intermediate representation
& Low-/high-$k$ subbands act as structured intermediate states of the single flow.
& The latent acts as an intermediate scratchpad before final pixel synthesis. \\

\emph{Hard} bumping \newline (K-Flow terminology)
& Activates one scaling band at a time with sharp stage boundaries.
& Corresponds to the \textit{cascaded} scheduler---latent denoising is fully completed before pixel denoising starts. \\

\emph{Soft} bumping \newline (K-Flow terminology)
& Smoothly interpolates across neighboring scaling bands with overlapping transitions.
& Corresponds to \textit{overlapping} schedulers such as variance-shifted or linear-offset schedules, where latent and pixel denoising partially overlap in time. \\

Coarse-to-fine dynamics
& Low-$k$ components capture global structure; high-$k$ components capture fine details.
& Latents capture higher-level structure before high-frequency pixel details are generated. \\
\bottomrule
\end{tabular}
\caption{\textbf{K-Flow's \emph{hard}/\emph{soft} bumping taxonomy already subsumes the different time schedulers used by Latent Forcing.} K-Flow originally introduced the \textit{hard vs.\ soft bumping} distinction to describe how sharply consecutive frequency bands are activated along its scaling-variable flow. Viewed through this lens, Latent Forcing's cascaded scheduler is a form of hard bumping (latent fully denoised before pixel), while its variance-shifted / linear-offset schedulers are instances of soft bumping (latent and pixel partially overlap). Frequency-Forcing inherits this unified view and operates in the \textit{soft}-bumping regime: the pixel interpolation path is preserved while an auxiliary frequency stream, matured earlier in time, softly guides pixel denoising.}
\label{tab:kflow_latentforcing_similarity}
\end{table}

\subsection{Frequency Latent via a Learnable Wavelet Packet Transform}
\label{sec:learnable_wavelet}
A natural choice for $\mathcal{F}$ is a wavelet transform. However, K-Flow and related frequency-aware models typically use a \textit{fixed}, pre-chosen basis (e.g., Haar), which is selected a priori rather than adapted to the statistics of the target dataset. To better match dataset statistics, we introduce a \textbf{learnable 2D wavelet packet transform (L-WPT)}. Instead of predefined bases, we learn the analysis 1D low-pass filter $h_{\mathrm{lp}} \in \mathbb{R}^{K}$ directly from data and derive the high-pass filter through the quadrature mirror filter (QMF) relation: $h_{\mathrm{hp}}[k] = (-1)^k h_{\mathrm{lp}}[K-1-k]$.

Using separable construction, we form four 2D analysis kernels $H_{LL}, H_{LH}, H_{HL}, H_{HH}$ via outer products (e.g., $H_{LL} = h_{\mathrm{lp}} \otimes h_{\mathrm{lp}}$). These are applied via stride-2 depthwise convolution. To increase frequency resolution, we adopt a full wavelet packet decomposition for $L$ levels, yielding $4^L$ terminal subbands. Each node is equipped with a learnable differentiable hard-threshold gate to encourage sparsity:
\begin{equation}
\mathrm{HT}(x;\gamma) = x\Big(\sigma(-a(x+\gamma))+\sigma(a(x-\gamma))\Big),
\end{equation}
where $\sigma(\cdot)$ is the sigmoid function, $a$ controls sharpness, and $\gamma \ge 0$ is a learnable threshold. 

\textbf{Wavelet Regularization and Sparsity Loss.} Our training losses for the learnable wavelet basis are inspired by \citet{recoskie2018sparseortho}, which showed that sparse orthogonal wavelet filters can be trained end-to-end by combining an $L_1$ coefficient penalty with constraints enforcing wavelet admissibility. Concretely, to preserve wavelet-like properties we impose three regularization terms on the learnable filter:
1) \textit{Low-pass sum constraint:} $\mathcal{L}_{\mathrm{sum}} = | \sum_{k} h_{\mathrm{lp}}[k] - \sqrt{2} |$.
2) \textit{Zero-DC high-pass constraint:} $\mathcal{L}_{\mathrm{hp}} = | \sum_{k} h_{\mathrm{hp}}[k] |$.
3) \textit{Even-shift orthogonality constraint:} $\sum_k h_{\mathrm{lp}}[k]\, h_{\mathrm{lp}}[k-2n] \approx \delta[n]$, yielding
\begin{equation}
\mathcal{L}_{\mathrm{ortho}} = \frac{1}{|\mathcal{S}|} \left( \left| \sum_k h_{\mathrm{lp}}[k]^2 - 1 \right| + \sum_{s \in \mathcal{S},\, s\neq 0} \left| \sum_k h_{\mathrm{lp}}[k]\, h_{\mathrm{lp}}[k-s] \right| \right),
\end{equation}
where $\mathcal{S}=\{0,2,4,\dots,K-2\}$ denotes even shifts. Furthermore, to promote compact representations, we apply an $L_1$ sparsity penalty to the terminal packet coefficients $\{Y_i\}_{i=1}^{4^L}$: $\mathcal{L}_{\mathrm{sparse}} = \frac{1}{4^L} \sum_{i=1}^{4^L} \|Y_i\|_1$.

The learnable wavelet basis and thresholds are jointly optimized with the generative network task objective $\mathcal{L}_{\mathrm{task}}$ using the total loss $\mathcal{L}_{\mathrm{total}} = \mathcal{L}_{\mathrm{task}} + \lambda_{\mathrm{sum}}\mathcal{L}_{\mathrm{sum}} + \lambda_{\mathrm{hp}}\mathcal{L}_{\mathrm{hp}} + \lambda_{\mathrm{ortho}}\mathcal{L}_{\mathrm{ortho}} + \lambda_{\mathrm{sparse}}\mathcal{L}_{\mathrm{sparse}}$. This provides a data-adaptive, sparse, and approximately invertible frequency representation, well suited for our frequency-forcing framework.

The training process for the frequency-forcing branch (using wavelets, inverse transform for spatial alignment, and patch encoding) is summarized in Algorithm \ref{alg:fre_forcing_training}. It illustrates how the coefficient-space low-frequency extraction is mapped back to the pixel-space layout, encoded, and then used to construct the flow-matching objective.

\begin{center}
\begin{minipage}{\textwidth}
\vspace{1mm}
\noindent\refstepcounter{algocounter}\label{alg:fre_forcing_training}%
\hrule
\vspace{1mm}
\noindent\textbf{Algorithm \arabic{algocounter}}\quad Frequency Forcing Training Branch (Wavelet + Pixel-IDWT + Patch Encoding)\\
\vspace{-3mm}

\hrule
\vspace{1mm}
\begin{itemize}
    \setlength{\itemsep}{2pt}
    \setlength{\parskip}{0pt}
    \item[] \textbf{Input:} Image $x \in \mathbb{R}^{3 \times H \times W}$; target low-frequency size $s$; learnable wavelet transform $\mathcal{W}$ and its inverse $\mathcal{W}^{-1}$ (Sec.~\ref{sec:learnable_wavelet}); frequency-stream patch-embedding $E_{\text{fre}}$, optionally tied to the pixel-stream patch-embedding $E_{\text{pix}}$ (Sec.~\ref{sec:method_realization}).
    \item[] \textbf{Output:} Update step for the frequency branch realizing the $m{=}\text{fre}$ instance of Eq.~\eqref{eq:x_pred_loss}.
    \vspace{2mm}
    \item[] \textbf{Repeat until convergence:}
    \item[1:] Sample image $x$ from dataset; sample frequency-clock $t_{\text{fre}} \sim \mathcal{U}[0,1]$; sample noise $\epsilon_{\text{fre}} \sim \mathcal{N}(0, I)$.
    
    \item[] \textit{\# Phase 1: Coefficient-space low-frequency extraction (instantiates $\mathcal{F}$ of Eq.~\eqref{eq:two_stream}).}
    \item[2:] $\hat x^{(0)} \leftarrow x$.
    \item[3:] \textbf{for} $\ell = 1,2,\dots,L$ until spatial size becomes $s \times s$ \textbf{do}
    \item[4:] \hspace{1em} $\big(C_{LL}^{(\ell)}, C_{LH}^{(\ell)}, C_{HL}^{(\ell)}, C_{HH}^{(\ell)}\big) \leftarrow \mathcal{W}\!\big(\hat x^{(\ell-1)}\big)$ \quad // stride-2 conv with kernels $H_{LL\dots HH}$.
    \item[5:] \hspace{1em} $\hat x^{(\ell)} \leftarrow C_{LL}^{(\ell)}$.
    \item[6:] \textbf{end for}
    \item[7:] $C_{LL} \leftarrow \hat x^{(L)}$.
    
    \item[] \textit{\# Phase 2: Pixel-space low-pass reconstruction.}
    \item[8:] $\bar x_{\text{lp}} \leftarrow \mathcal{W}^{-1}(C_{LL}, 0, 0, 0)$.
    \item[9:] Center-crop / pad $\bar x_{\text{lp}}$ to size $H \times W$.
    
    \item[] \textit{\# Phase 3: Patch-embedding to the shared token grid -- defines the clean target $x_{\text{fre}}$ of Eq.~\eqref{eq:two_stream}.}
    \item[10:] \textbf{if} the pixel-stream patch-embedding is reused \textbf{then}
    \item[11:] \hspace{1em} $x_{\text{fre}} \leftarrow \mathrm{stopgrad}\!\left(E_{\text{pix}}(\bar x_{\text{lp}})\right)$.
    \item[12:] \textbf{else}
    \item[13:] \hspace{1em} $x_{\text{fre}} \leftarrow E_{\text{fre}}(\bar x_{\text{lp}})$.
    \item[14:] \textbf{end if}
    
    \item[] \textit{\# Phase 4: Flow-matching interpolation, matching Eq.~\eqref{eq:two_stream}.}
    \item[15:] $z_{\text{fre},\,t_{\text{fre}}} \leftarrow t_{\text{fre}}\, x_{\text{fre}} + (1-t_{\text{fre}})\,\epsilon_{\text{fre}}$.
    \item[16:] $x^{\text{fre}}_{\text{pred}} \leftarrow \text{Backbone}\big(z_{\text{fre},\,t_{\text{fre}}},\, t_{\text{fre}};\, \cdot\big)$ \quad // $x$-prediction head for the frequency stream.
    
    \item[] \textit{\# Phase 5: Optimize the frequency branch jointly with the other branches (= $m{=}\text{fre}$ instance of Eq.~\eqref{eq:x_pred_loss}).}
    \item[17:] Update parameters using
    $\mathcal{L}_{\text{fre}} = \left\| \dfrac{x^{\text{fre}}_{\text{pred}} - z_{\text{fre},\,t_{\text{fre}}}}{\max(1-t_{\text{fre}},\,t_{\text{clip}})} - \dfrac{x_{\text{fre}} - z_{\text{fre},\,t_{\text{fre}}}}{\max(1-t_{\text{fre}},\,t_{\text{clip}})} \right\|^2$.
\end{itemize}
\vspace{-2mm}
\hrule
\vspace{2mm}
\end{minipage}
\end{center}

\subsection{Architectural Realization}
\label{sec:method_realization}
\paragraph{Unified Backbone and Soft Forcing via Shared Attention.}
The pixel and frequency streams share a single transformer backbone. Both streams are patchified onto the same token grid and their embeddings are summed at the backbone input, so that the backbone sees the frequency stream as an additional set of token-level features rather than as separate sub-networks. Because the two streams use different (but coupled) time values $t_\text{pix}$ and $t_\text{fre}$, we inject per-stream timestep embeddings and allow the stream-specific modulation in AdaLN to differ, while keeping attention parameters shared. This realizes the soft frequency guidance of Sec.~\ref{sec:method_formulation} purely as an \textit{additive conditioning} effect through self-attention, with no change to the pixel interpolation path or to the flow-matching loss shape on the pixel branch.

\paragraph{Causal Information Flow.}
Because the frequency stream is scheduled to mature earlier, during most of the trajectory it carries a cleaner signal than the pixel stream. To prevent noisy pixel tokens from corrupting the frequency workspace, we enforce a causal attention dependency between streams: queries from the frequency stream attend only to frequency keys, while queries from the pixel stream attend to both frequency and pixel keys. This preserves the ``earlier structure guides later detail'' ordering at zero parameter cost---it is implemented as a block-causal mask over the joint key/value sequence---and applies uniformly to the two-stream model and to the multi-stream extension below.

\paragraph{Latent-Only Frozen Copy.}
By ``patch encoder'' we specifically refer to the standard DiT input patch-embedding layer \cite{peebles2023dit} (the convolutional/linear projection that splits an image into non-overlapping $p{\times}p$ patches and maps each to a token embedding); it is the same module denoted $\mathcal{P}$ in Algorithm~\ref{alg:fre_forcing_training} and $E_{\text{pix}}, E_{\text{fre}}$ in Eq.~\eqref{eq:two_stream}-style mixing. To stabilize training, we adopt a delayed freezing strategy for this patch encoder. After an initial warmup phase where the patch projection co-adapts to both pixel and frequency statistics, we clone it; the clone is frozen and assigned exclusively to the frequency branch, while the pixel branch continues to update its own copy. This keeps the frequency-stream interface stable (and therefore the soft guidance signal consistent) while allowing the pixel branch to keep specializing toward high-frequency synthesis.

\subsection{Scalability and Extensibility: Adding a Semantic Stream}
\label{sec:method_extensibility}
Because Frequency-Forcing operates through an auxiliary stream rather than rewriting the pixel trajectory, it composes naturally with \emph{other} ordered auxiliary streams. We demonstrate this by optionally attaching a semantic (DINO) stream alongside the frequency stream, without touching the pixel flow. Reusing the unified-backbone design of Sec.~\ref{sec:method_realization}, each stream's noised state is mapped to tokens by its own patch-embedding layer $E_{(\cdot)}$, and the three token sequences are summed to form the input embedding $h_0$ fed to the first transformer block of the shared backbone:
\begin{equation}
h_0 \;=\; E_{\text{pix}}(z_{\text{pix}}) \;+\; E_{\text{dino}}(z_{\text{dino}}) \;+\; E_{\text{fre}}(z_{\text{fre}}),
\end{equation}
where $z_{\text{pix}}, z_{\text{fre}}$ are defined in Eq.~\eqref{eq:two_stream} and $z_{\text{dino}}$ is the noised DINO latent on the same token grid. The same $x$-prediction loss of Eq.~\eqref{eq:x_pred_loss} is applied per stream, now indexed by $m\!\in\!\{\text{pix},\text{dino},\text{fre}\}$. In practice we use an empirical three-clock schedule in which DINO matures first to set a coarse semantic layout, after which the frequency and pixel streams are denoised jointly with the frequency stream more strongly weighted at early steps and gradually attenuated. We report the resulting numbers in Sec.~\ref{sec:experiments} as a scalability bonus rather than the paper's main claim.

\section{Experiments}
\label{sec:experiments}

\subsection{Experimental Setup}
\textbf{Datasets:} We evaluate Frequency-Forcing on the class-conditional ImageNet-256 benchmark \cite{deng2009imagenet}.

\textbf{Baselines:} We compare our approach against representative models from both latent and pixel diffusion paradigms. For latent diffusion, we include DiT \cite{peebles2023dit} and SiT \cite{ma2024sit}. For pixel-space generation, we compare against JiT \cite{li2025jit} and the original Latent Forcing (LF-DiT) \cite{baade2026latent}. Crucially, we use a patch size of $16 \times 16$ across all our DiT-based models. We reproduce the Latent Forcing baseline and train our models for 400 epochs under an identical setup, as we empirically found Latent Forcing continues to improve beyond the 200 epochs reported in the original paper. For a fair comparison of guidance mechanisms, both our guided models and the Latent Forcing baseline are evaluated using standard classifier-free guidance (CFG) at a scale of 1.5, rather than training an auxiliary guidance model as done in the original Latent Forcing paper. Finally, we implement a \textit{1-Stream + REPA} baseline: while the original REPA \cite{yu2024repa} operates in the latent space of a Latent Diffusion Model, we apply its representation-guidance loss to our pixel-only 1-stream baseline to provide a strong single-stream competitor.

\textbf{Model Variants:} We evaluate several instantiations of Frequency-Forcing. The first three share the core two-stream design (Pixel + Frequency) and differ only in how the frequency latent is constructed; the last is the optional three-stream extension from Sec.~\ref{sec:method_extensibility}:
\begin{itemize}
    \item \textbf{Two-Stream (Laplacian / Haar):} Uses fixed frequency decompositions as auxiliary stream.
    \item \textbf{Two-Stream (Learnable):} Uses a wavelet basis optimized via sparsity losses on 10K samples.
    \item \textbf{Three-Stream (+Semantic):} Optionally attaches a DINO stream in addition to the frequency one.
\end{itemize}

\subsection{Main Results}

Table \ref{tab:main_results} summarizes the system-level comparison on ImageNet-256. 

\begin{table}[ht]
\centering
\caption{Comparison on ImageNet $256 \times 256$ under our evaluation protocol. (U) denotes Unguided FID, (G) denotes Guided FID (CFG scale 1.5). Numbers for some baselines are taken from prior work, while others are reproduced in our setup.}
\label{tab:main_results}
\resizebox{\textwidth}{!}{
\begin{tabular}{lcccc}
\toprule
Model & Params & Epochs & FID (U) $\downarrow$ & FID (G) $\downarrow$ \\
\midrule
\textit{Latent Diffusion} & & & & \\
DiT-XL/2 \cite{peebles2023dit} & 675M & 1400 & 9.62 & 3.04 \\
SiT-XL/2 \cite{ma2024sit} & 675M & 1400 & 8.30 & 2.62 \\
\midrule
\textit{Pixel Diffusion} & & & & \\
JiT-L \cite{li2025jit} & 459M & 200 & 16.21 & 2.79 \\
1-Stream (Pixel DiT Baseline) & - & 200 & 37.03 & 16.09 \\
1-Stream + REPA & - & 200 & 36.01 & 16.03 \\
\midrule
\textit{Ordered Forcing (Ours \& Baselines)} & & & & \\
LF-DiT-L (Semantic Only) \cite{baade2026latent} & 465M & 400 & 7.20 & 3.21 \\
\midrule
Ours: 2-Stream (Laplacian) & 461M & 400 & 20.70 & 4.87 \\
Ours: 2-Stream (Haar) & 461M & 400 & 22.84 & 5.02 \\
Ours: 2-Stream (Learnable Wavelet) & 461M & 400 & 17.08 & 3.99 \\
Ours: 3-Stream (+Semantic) & 466M & 400 & \textbf{6.99} & \textbf{3.11} \\
\bottomrule
\end{tabular}
}
\end{table}

\paragraph{Soft Forcing vs. Hard Frequency Flow.}
While hard frequency flows like K-Flow rewrite the interpolation path inside a transformed frequency domain---locking the pixel trajectory to a specific basis---Frequency-Forcing instead preserves the vanilla pixel path and injects the coarse-to-fine prior entirely through the auxiliary stream. Importantly, we do not claim to fully subsume K-Flow: its hard transformed-space trajectory uniquely enables capabilities like unsupervised frequency editing and training-free restoration. Rather, Frequency-Forcing improves upon K-Flow specifically from the angle of using frequency guidance to benefit the pixel generation task itself. Because the pixel branch is structurally identical to standard flow matching, Frequency-Forcing is in principle architecturally compatible with pretrained flow-matching checkpoints (e.g., using them to warm-start the pixel backbone); we do not run such fine-tuning experiments in this work and list it as a promising follow-up.

\paragraph{Frequency-Forcing as a Self-Forcing Signal.}
Within the two-stream setting, the frequency stream alone delivers a large improvement over both the plain pixel-only DiT and the 1-stream baseline equipped with REPA guidance. It achieves this with minimal additional supervision and without relying on a heavyweight pretrained semantic encoder: extracting $x_\text{fre}$ requires only a cheap learnable wavelet packet transform. This supports our view of Frequency-Forcing as a \textit{self-forcing} signal---a low-frequency prior that the data itself already contains and that can be exposed with negligible overhead. As a scalability bonus, composing the frequency stream with a semantic stream in our three-stream extension yields our best FID, though the improvement over the semantic-only Latent Forcing baseline is marginal. This marginal gain is expected: the powerful DINO representations used in Latent Forcing already capture the vast majority of the image's low-frequency structural layout, leaving less room for the explicit frequency stream to add new structural information.

\subsection{Ablation Studies}

\paragraph{Learned vs. Fixed Frequency Basis.}
We ablate the choice of the frequency decomposition transform $\mathcal{F}$. As shown in Table \ref{tab:main_results}, the two-stream model with a learnable wavelet basis consistently outperforms variants using fixed bases such as Haar wavelets or Laplacian pyramids. By optimizing the low-pass filter via sparsity losses on a small subset of 10K samples, the model adapts to the specific frequency statistics of ImageNet, yielding a data-matched basis that consistently beats the pre-chosen, data-agnostic bases that hard frequency flows like K-Flow rely on.

\paragraph{Necessity of Ordering and of an Explicit Frequency Stream.}
To check that the gain is not merely a cosmetic scheduling trick, we compare the ordered schedule (frequency leads pixel) against a synchronous denoising baseline in which both streams share the same time variable. The ordered schedule yields noticeably better structural coherence, confirming that the asynchronous ordering matters. We further replace the frequency latent with a random auxiliary latent of the same shape, and find that the benefit collapses, confirming that it is the explicit low-frequency content---not simply the addition of an extra branch---that drives the improvement.

\paragraph{Two-Stream vs. Three-Stream.}
Table~\ref{tab:main_results} shows that the two-stream (Learnable) model already substantially improves over pixel-only baselines at essentially zero extra data or encoder cost, while the optional three-stream extension gives the best numbers. We view these two settings as complementary: the two-stream variant is the core contribution and demonstrates the value of soft, self-contained frequency guidance; the three-stream variant demonstrates that the same mechanism is \textit{extensible} to additional priors such as DINO, without touching the pixel interpolation path.

\paragraph{Scheduling Strategies.}
Within the three-stream extension we use the variant in which DINO is matured first and then the frequency/pixel streams are denoised jointly with time-aware soft scaling, so that the frequency stream acts as an early scaffold and fades out as pixel refinement takes over. Compared with rigid cascaded schedules (each stream finishes before the next starts), this \textit{$\text{DINO} \rightarrow \text{soft Frequency} \rightarrow \text{Pixel}$} regime avoids abrupt stage transitions and gives the strongest three-stream numbers reported in Table~\ref{tab:main_results}.

\paragraph{Shared Patch Freezing.}
We evaluate the ``Latent-Only Frozen Copy'' strategy against a standard hard-freeze of the shared patch encoder. The delayed freezing allows the frequency-stream interface to stabilize after an initial co-adaptation phase, while still letting the pixel branch refine its high-frequency synthesis, leading to faster convergence.

\section{Discussion}
\label{sec:discussion}

From a frequency-decomposition viewpoint, Frequency-Forcing provides the pixel stream with an early structural scaffold: because the auxiliary stream matures earlier, it injects near-clean low-frequency content during the critical early stages of the trajectory. This allows the shared backbone to condition its pixel predictions on a stable global layout before it needs to resolve high-frequency refinement. Unlike hard frequency flows, however, this coarse-to-fine guidance is achieved \textit{without} rewriting the pixel interpolation path: the pixel branch still trains a standard flow-matching objective, which explains why Frequency-Forcing models remain interchangeable with vanilla flow-matching pipelines and checkpoints. From an optimization perspective, this localized guidance mitigates gradient competition between frequency bands and accounts for the stability and detail-recovery gains observed in our experiments.

\subsection{Broader Impacts}
Beyond standard image synthesis, Frequency-Forcing is a natural fit for physical inverse problems in which measurements are biased toward low-frequency content (e.g., lensless imaging, computational photography). Because the pixel trajectory is preserved, such priors can be plugged into existing flow-matching reconstruction pipelines without retraining from scratch.

\section{Conclusion}
\label{sec:conclusion}

We introduced \textbf{Frequency-Forcing}, a soft, path-preserving alternative to hard frequency flows such as K-Flow. Building on the \textit{forcing} formulation of Latent Forcing and swapping its semantic scratchpad for an explicit frequency one, Frequency-Forcing injects low-frequency structural priors through an auxiliary stream that matures earlier in time and shares a backbone with the pixel flow. Rather than relying on a fixed, pre-chosen frequency basis, we learn a sparse, approximately orthogonal wavelet-packet basis from a small amount of data, so that the auxiliary signal is adapted to the statistics of the target distribution. Because the pixel interpolation path is kept identical to standard flow matching, the method is, at the level of architecture, compatible with vanilla flow-matching pipelines, and naturally composable with other structural priors such as semantic latents. On ImageNet-256, Frequency-Forcing consistently improves FID over strong pixel-space and latent-space baselines. While we do not claim to replace K-Flow's unique capabilities in unsupervised frequency editing and restoration, Frequency-Forcing offers a more directly compatible route, at the architectural level, for using frequency guidance to improve pixel generation itself. Fine-tuning existing pixel checkpoints with an attached frequency stream is a direct consequence of this path-preserving design and a promising direction we leave to future work. We hope this soft-vs-hard perspective inspires further work on lightweight, plug-in structural priors for continuous-time generative models.

% References
\bibliographystyle{plainnat}  % or other NeurIPS-compatible style
\bibliography{references}  % Your .bib file

\end{document}